# A Pattern-Hierarchy Classifier for Reduced Teaching

Kieran Greer, Distributed Computing Systems, Belfast, UK.
http://distributedcomputingsystems.co.uk
Version 1.2

*Abstract -* This paper describes a design that can be used for Explainable AI. The lower level is a nested ensemble of patterns created by self-organisation. The upper level is a hierarchical tree, where nodes are linked through individual concepts, so there is a transition from mixed ensemble masses to specific categories. Lower-level pattern ensembles are learned in an unsupervsised manner and then split into branches when it is clear that the category has changed. Links between the two levels define that the concepts are learned and missing links define that they are guessed only. This paper proposes some new clustering algorithms for producing the pattern ensembles, that are themselves an ensemble which converges through aggregations. Multiple solutions are also combined, to make the final result more robust. One measure of success is how coherent these ensembles are, which means that every data row in the cluster belongs to the same category. The total number of clusters is also important and the teaching phase can correct the ensemble estimates with respect to both of these. A teaching phase would then help the classifier to learn the true category for each input row. During this phase, any classifier can learn or infer correct classifications from some other classifier's knowledge, thereby reducing the required number of presentations. As the information is added, cross-referencing between the two structures allows it to be used more widely, where a unique structure can build up that would not be possible by either method separately.

## 1   Introduction

This paper describes a design that can be used for Explainable AI. It can probably be integrated into an existing cognitive model [14] at the boundary between the middle and upper levels, where the knowledge-based and the experienced-based systems share information. The lower level is a nested ensemble of patterns created by self-organisation. The upper level is a hierarchical tree, where each end node represents an individual concept, so there is a transition from mixed ensemble masses to specific categories. Lower-level pattern ensembles are learned in an unsupervsised manner and then split into branches when it is clear that the category has changed. Links between the two levels define that the concepts are learned and missing links define that they are guessed only. This paper proposes some new clustering algorithms for producing the pattern ensembles, that are themselves an ensemble which converges through aggregations. Multiple solutions are also combined, to make the final result more robust. One measure of success is how coherent these ensembles are, which means that every data row in the cluster belongs to the same category. The total number of clusters is also important and the teaching phase can correct the ensemble estimates with respect to both of these. A teaching phase would then help the classifier to learn the true category for each input row. During this phase, any classifier can learn or infer correct classifications from some other classifier's knowledge, thereby reducing the required number of presentations. This would lead to each category aggregating several associated clusters, where sub-clusters split from the ensemble and then link to the category node. As the information is added, cross-referencing between the two structures allows it to be used more widely.

With this process, a unique structure can build up that would not be possible by either method alone. The two structures link-up the same data row between the ensemble and the tree, and tree end nodes represent the learned knowledge. If a pattern sub-cluster becomes associated with two or more categories, that sub-cluster is separated, but only needs to recognise the difference in the row sets that belong to its base classifier, not the whole dataset. The discrimination problem is therefore made simpler by reducing the problem size. There is also a lot of cross-referencing between the self-organised clusters and the taught tree and this would help the classifier to learn more quickly and to share partial results. The algorithms in this paper mostly use processes and equations that the author has used previously, but it is more important to understand the broad algorithm and underlying theory, because a lot of the functions could probably be replaced by other ones.

The rest of this paper is organised as follows: section 2 describes some related work. Section 3 describes the unsupervised clustering theory, while section 4 describes two clustering algorithms that have been tried. Section 5 then introduces the supervised clustering theory with the teaching. Section 6 gives some test results, while section 7 gives some conclusions on the work.

## 2   Related Work

A relatively recent AI topic is Explainable AI (XAI) [9]. With this, the AI system is able to give an explanation, in human terms, of how it came to a decision. This is intended to increase trust in the system that is no longer a black box, but can be more transparent. It would also allow humans to interact with the system more easily because it will have to share a common language for the explanation. DARPA [8] consider this to be the next stage in AI, especially with regard to autonomous systems that may take actions on their own. Concerning this paper, there is a small amount of feedback available from the new structure that could be used to allow for more intelligent interaction by a human operator. Other AI models have been developed based on similar designs. The Adaptive Resonance Theory neural network [6][7], for example, has an architectural similarity. There are different variants, but it was designed with an unsupervised bottom layer that would cluster the input and try to match it with a static set of categories (memory) in the upper layer. This could lead to new categories being created if there was a mismatch and the weight sets would be updated each time, to re-align themselves with the current cluster sets. The network was later found to suffer from a statistical property that meant the order in which the data was presented would affect how it was clustered. A relatively new version called TopoART [19] is able to address this problem, as the shapes of the clusters do not depend on the order of creation of the associated categories. The Frequency Grid algorithm [15] that is used later, also suffers from this problem, where an improvement is suggested in section 4.2.1.

With regard to the unsupervised part, the author has published some related papers. The ensemble-hierarchy model has close relations to one suggested for the cognitive model [14]-[16]. The Self-Organising Map [18] is obviously of interest, or SOM with extensions [5], as are the clustering algorithms that consider sets of closest nodes, such as DBSCAN [10], kNN or k-Means [20]. A difference with these algorithms is that the new method considers not only the points nearest to the node in question, but the points nearest to its nearest nodes as well. A fully-connected architecture has always been suggested for a biological model, see for example [2]. Random Forests [3][1] are another ensemble method that are used with Decision Trees [4]. While Decision Trees branch on attributes and not category, the clustering process is very similar. The dataset is split into *n* different sets, each with maybe 60-80% of the original dataset. Each variation is trained on a Decision Tree and the results are aggregated together. The Random Forest is the training process that uses multiple variations of the dataset and also the aggregation process afterwards.

The paper [13] makes some interesting comments about Boolean Factor Analysis that would relate to this ensemble-hierarchy and may therefore be earlier related work. Their Hopfield-modified network takes the input signal vector and factors it into a low-level signal space of relations or clusters. The low-level factors would represent the first clustering stage. One idea is to further self-organise based on distinct features, as well as closest distances. Columnar characteristics can therefore become important and decisions can be taken, maybe with some judgement on related features. At the heart of Deep Learning [17] is the idea of learning an image in discrete parts. Each smaller part is an easier task and the next level can then combine the smaller parts until the whole image is learned. It might be interesting to compare the branching with something like this, because it also reduces the problem complexity.

## 3    Unsupervised Clustering Theory

Self-organisation is more often used to extract patterns from data, than to learn known categories and does this using some type of distance or similarity measurement. The self-organising process relies on some basic theories as follows: The process starts by associating every data row with the row it is closest to, according to some measure, such as Euclidean Distance. If each row is then clustered with its closest row, this should actually lead to natural breaks in the data that lead to a set of natural clusters. It is very likely that there will be more clusters than actual categories in the dataset and so each actual category will be represented by several clusters. However, if each cluster is considered in isolation, it will also be found to have sub-clusters that can be recognised through the same closest link mechanism. These sub-clusters are only obvious when the larger enclosing pattern is removed and the cluster is considered by itself. The sub-clusters might then be helpful, because they can isolate data rows that do not really belong together. Clustering using centroids has to consider average values, where it cannot skew weight values to obtain a desired result. Therefore, data from different categories can easily get mixed together. A re-clustering phase would then be able to move isolated data rows to other clusters and add data rows that may belong to the main category of this cluster. Through this method, the cluster may become a centre of attraction for the category it represents and its centroid will become more accurate, as more and more data rows for the same category are added.

While that is the theory, it does not work out quite so well in practice. One big problem with self-organisation is the fact that it has to choose the centre of the data that it is clustering. The algorithm does not know what the actual category is and so it cannot directly discriminate. Taking any sub-clusters too literally is probably dangerous as well, because the averaged values rely on there being distance consistency across the global cluster, which does not have to be the case. This is OK if there are few categories and the data is well-balanced, but the self-organising mechanism cannot learn any inherent skew in the dataset. A supervised approach, on the other hand, is able to adjust its discrimination lines, because it can be told directly about a particular error and so it can then adjust a weight set based on this. The teaching phase is therefore intended to make the self-organised patterns more accurate. It is postulated that because some of the classification has been learned and can be used as a sound basis, the teaching phase can help to build up a more accurate picture of the whole data set with fewer presentations.

## 4   Clustering Algorithms

Two clustering algorithms have been tried for this problem. A first attempt was based more on node distances and is not the current algorithm of interest, whereas a second attempt, based more on the Frequency Grid, is currently the algorithm of interest.

### 4.1 Distance-Based

This first attempt uses a k-means clustering to create the large clusters and then a frequency grid inside of each cluster to split them again, with the intention of providing some robustness through shifting the centroid centres each time. The frequency grid is equivalent to clustering based on popular count associations. The algorithm could be accurate in some cases but it produced too many clusters to be practical. This failing is shown in Table 1 of section 6. The self-organisation phase would cluster based on closest distance, but it would also try to ceate the largest and most coherent cluster sets possible. Algorithm 1 gives an example for this type of clustering:

Algorithm 1. Closest-Distances Example
---

1. Link each data row with the row it is closest to, according to some measure.
2. Create clusters by placing all data rows that are linked together into a cluster.
3. For each cluster:
    a. Use a Frequency Grid to do a count of the rows any other row is closest to.
    b. Use the grid to create sub-clusters in the cluster.
    c. Special cases include a sub-cluster with only 1 entry, or single column features.
4. Create branches in each base classifier for each sub-cluster part and create a centroid for each sub-cluster. Also add a new sub-cluster for any additional rows.
5. Try to combine any of the base clusters as follows:
    a. Determine an average distance $\bar{u}$ between the sub-clusters in the cluster.
    b. Determine a distance $x$ between two clusters.
    c. If the distance $x$ is less than the average sub-cluster distance $\bar{u}$, then combine the two clusters.
6. Re-calculate the centroids for each cluster and sub-cluster.
7. Take each data row in turn again and add it to the cluster whose centroid it is now closest to. Go to step 3.
8. The process can stop when data rows are not moved or the total number of clusters does not change.

### 4.2 Distance and Frequency Grid

The algorithm of section 4.1 could be accurate in some cases, but it is also a bit messy. Relying on node distances inside of clusters is not very reliable. The average row position and distance can change across a single cluster, for example. The algorithm of this section is a lot cleaner but also a bit simplistic and so extensions to it are also suggested. This second algorithm uses a brain-inspired idea of full linking between the nodes, to find a better closest match. A difference with something like DBScan, kNN or k-means is that this algorithm considers not only the points nearest to the node in question, but the points nearest to its nearest nodes as well. Each clustering phase measures the k-nearest clusters to every other cluster, where a cluster is a single node to start with. But instead of just considering the node's k-clusters, it aggregates all of the k-clusters for the nodes closest to the node in question. The intention is to produce a more robust association count, which can consider that while a node may be closer, if it is really part of a different category, it will have other associations that the rest of the k-cluster nodes do not have. Filtering over a combined and cross-referenced list for several mini clusters, would therefore help to remove this as noise. The algorithm for the brain-inspired with frequency grid clustering is described in Algorithm 2.

Algorithm 2. Single Pass Algorithm

1. Set each node, representing a single data row, to be a separate cluster.
2. Create a new layer of clusters using Algorithm 3.
3. Set the new layer as the next layer to cluster and check the stopping criteria.
   3.1. This can include a closest match with a preferred number of clusters.
4. If stopping criteria not met, then Go to step 2.

Algorithm 3. Brain-Inspired Frequency Grid

1. Measure the distances between all of the clusters and for each cluster, store the closest *k* other clusters.
2. Each cluster then is associated with *k* other clusters and they are also each associated with *k* other clusters.
3. For each cluster:
   3.1. Do a count over all of the associated cluster names, to find the most popular *k* clusters overall.
   3.2. Store this set as the local cluster set for the cluster node.
4. Convert each cluster set into an event, or input data line for the frequency grid.
   4.1. Train the frequency grid with the input data lines and it will produce another set of clusters, based on most popular association counts.

- 4.2. Create a new layer in the model that groups all nodes (clusters) suggested by the frequency grid together.
- 4.3. Let each new cluster be represented by its centroid value, or averaged data row.

### 4.2.1 Ensemble Improvement

The problem with the frequency grid is that it is sensitive to the order in which nodes are processed. This includes simply the order that they occur in a layer of nodes. If the dataset is randomly ordered therefore, it will have an affect on the result and so it may be better to produce several results that are related to each other and aggregate over those, so that the more commonly occurring parts are kept and the more-noisy parts are removed.

There are at least two solutions to the problem of random ordering. One option would be to make the set membership fuzzy, where the frequency grid would allow a node to be a member of all clusters that have an equal association count with it. But that drastically changes the nature of the frequency grid. It is supposed to have that type of relationship for between-cluster links, but not for the full membership of more than one cluster. It could force those clusters to merge, when they would lose some meaning. The second option is to use an ensemble of solutions, in the style of random forests. For this problem, the random solutions stem from the same base and so they have the same root set of values. This is simply something that might be statistically significant. If several solutions result, the more commonly occurring parts will remain and the noisy or less commonly occurring parts will receive lower statistical counts and be removed. The batch process that uses the ensemble algorithm is described in Algorithm 4.

Algorithm 4. Batch process for averaged test results

1. While (run ensemble test)
    a. For (i = 0 to test runs)
        i. Generate a randomised data ordering.
        ii. Use Algorithm 3 to produce a cluster set.
        iii. Save the cluster set as a solution set.
    b. Process all solution sets together, using the ensemble method of Algorithm 5.

    i. This produces something like (tr*tr*n) new solutions from the multiple single-pass algorithm runs.
  c. Convert the new solutions to cluster lists of node names.
  d. Use cluster lists as the input to another frequency grid. Only one final stage, so don't randomise.
  e. Convert frequency grid clusters to a new cluster list of node names.
  f. Create new layer and nodes from that ordering and cluster using Algorithm 3.
  g. Save these clusters as a solution.
2. If stopping criterion met, then stop.
3. Average the results to give the final result set.

Algorithm 5. Ensemble algorithm

1. For (i= 0 to phases - *helps to reduce the number of clusters*)
   a. Convert last solution set to cluster lists of node names.
   b. For (j = 0 to test runs)
      i. Randomise the first cluster list ordering.
      ii. Use as input to a frequency grid.
      iii. Convert frequency grid clusters to second cluster list of node names.
      iv. For (k = 0 to test runs)
         1. Randomise the second cluster list ordering.
         2. Create new layer and nodes from that ordering.
         3. Cluster the new layer using Algorithm 3.
         4. Save the result as a solution set for the next phase.

### 4.2.2 Consolidate through Hierarchy

In fact, there is a further opportunity to use a clustering process at the very end of the ensemble tests. If each ensemble run produces a result, then instead of averaging the result scores, the result solutions can be stored and used as the input node sets for a final clustering process. This introduces the idea of hierarchical clustering that is beyond the hierarchy created by the frequency grid convergence. It is more like the whole process being repeated, using the result of the first stage as the input.

## 5 Supervised Teaching Theory

New to the system would be a second supervised stage that would teach the correct category and help to explain the classification decisions. When a value for an actual category is learned, this information might also be used to measure a confidence that the classification is correct. For example, if links between the self-oganised cluster and the taught cluster do not exist, then there is a high probability of a guess, but if links are present, then the information of the self-organised cluster can be used with confidence. Figure 1 is a graphic that describes some of the processes that would be involved.

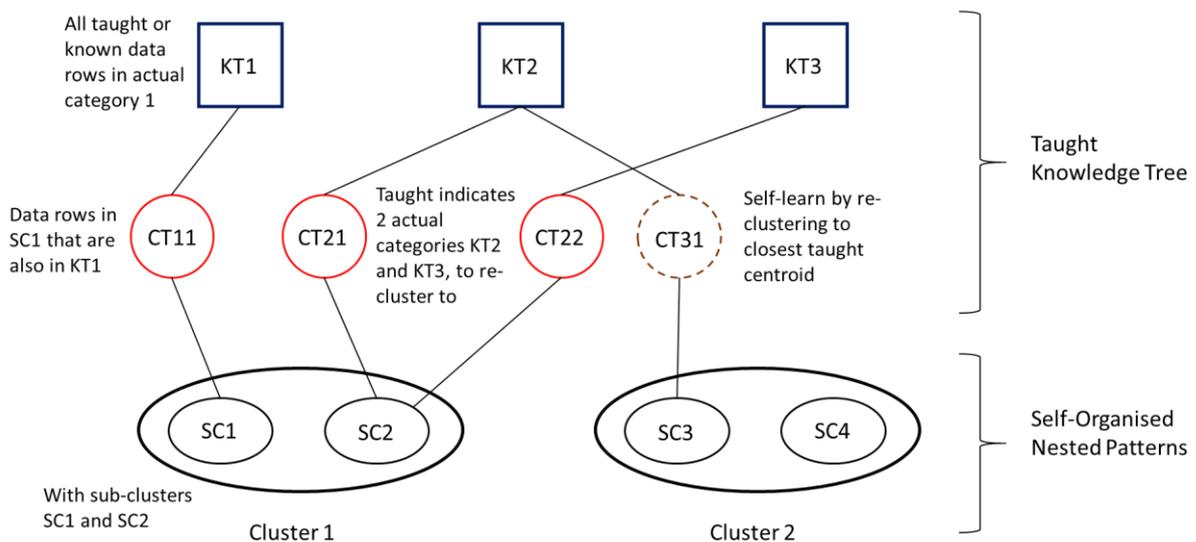

Figure 1. Graphic of the possible interactions between the two cluster structures.

With the supervised phase, the classifier is allowed to ask for the actual category of a random data row. The idea is that in the real world we may make some assumptions based on what we can determine, but we would also know that they are guesses. We would wait for proof before considering them to be true and we would then use the 'known' knowledge to correct any of the related assumptions. The classifier can therefore ask for some random proof and use it to correctly classify that data row in the related cluster. The proof is also added to the knowledge tree from where it can be used by any of the cluster groups. It can also return a true centroid value for the specific category. Any classifier then has the option of using this knowledge to update their own data associations, for example. Also, through this process, actual category values can be assigned to what was previously only classified as a coherent group.

As more rows are learned, the category node in the tree can become more accurate and there may be a constant ripple effect of updating the centroids and re-assigning the data rows without major structural changes. However, more major merging or creation of new clusters is also required. For example, at some stage, the classifier will receive information that means a cluster is now associated with more than 1 category. Nodes can then be added between the split clusters and the category, to link them, where these inter-nodes are only required to recognise a difference between the row sets for the sub-cluster. With cross-referencing, the centroid information can also be shared between the clusters and any of the node sets may grow or reduce in size as data rows get re-assigned.

## 6  Implementation and Testing

It has only been possible to test the self-organising structure so far. A computer program has been written in the C# .Net language and it tested some benchmark datasets that can all be found in the UCI Machine Learning repository [21]. Two tests have been carried out, one for the algorithm of section 4.1 and one for the algorithm of section 4.2. No information about the clusters was given to the program, except for the desired number as a stopping criterion. The clusters were generated internally by the program, resulting in sets of nodes inside of each cluster that would hopefully be coherent with each other, meaning that they would all belong to the same actual category. It was then possible to calculate the error for those as follows:

1. For every sub-cluster, retrieve from the dataset, the category for each row.
2. Remove the set of rows with the largest count for a single category.
3. The coherence error is then the number of rows left.

So, for example, if a cluster set contains data rows for categories as follows: A, A, A, B, B, then there would be 2 incorrect nodes. If the dataset actual categories were: A, A, A, B, B, C, C, then the error would be 4. The tests in section 6.1 are really only a marker and a more complete set of tests is described in section 6.2.

## 6.1  Distance-Based

As a marker to compare with, tests were firstly carried out on the algorithm of section 4.1. If the data was already well separated and the number of actual categories was low, then the self-organising process could realise the original categories by itself, but a stopping criterion, or knowing when it had converged might be problematic. This was the case for the Iris [11] and the Wine [12] and Zoo [22] datasets. A lot of other datasets showed that the self-organising structure cannot perform well enough by itself. This was also found to be the case in [5] who used variants of the SOM to successfully cluster the Iris data but could not cluster the Abalone dataset, for example. It was also a characteristic of the process that in cases when incoherent data was higher, it might start with a smaller number of clusters, but by trying to reduce this, the number would in fact increase. So, by trying to move data rows from the first-assigned cluster would in fact fragment the clusters more. Other factors such as re-combining clusters, can also really increase the error.

| Dataset | Incoherent | S-O Clusters | Actual Clusters |
|---|---|---|---|
| Iris | 2 of 150 | 30 | 3 |
| Wine | 4 of 178 | 18 | 3 |
| Zoo | 7 of 101 | 18 | 7 |

Table 1. Example of self-organised coherence. 'Incoherent' shows how many data rows were not coherent, or of the same actual category, as the rest of their cluster. 'S-O Clusters' shows how many separate clusters were created. 'Actual Clusters' gives the correct number.

**6.2   Distance-Based and Fequency Grid**

This section gives test results for the algorithm of section 4.2, which could probably be used in the real world. For this test, the preferred number of clusters would be entered and the algorithm would be run and stopped at the number of clusters just above or equal to the preferred number. The algorithm would initially decide the cluster sets using the closest-node associations and then reduce the number using the frequency grid. Comparisons were made between ordered and random datasets, and the single pass algorithm or the ensemble version. The only configuration for the ensemble was the number of runs inside the ensemble, set at 10 and the number of closest-node associations, set at 5.

Using an ensemble is still a heuristic process and the result of each ensemble search can be different. The program therefore ran a number of ensemble tests for each dataset configuration and then averaged the results to get the final totals. With the ensemble testing, there would be a maximum of 50 separate test runs for each dataset configuration. Each run would produce an averaged number of clusters and percentage of correct associations, for the clustering process. Those 50 results would then be averaged to produce a final result for the test. The test results are listed in Table 2. As a comparison, the first column gives the result using the closest-nodes clustering only, which is the number of clusters followed by the accuracy percentage. A single run for the ordered dataset is given in the second column. This is followed by a single run for the randomised dataset and then the fourth column gives the ensemble result. A fifth column then gives the result if a final clustering stage is also applied, maybe as part of a hierarchy. There are two versions of the frequency grid algorithm, indicated by 'V1' or 'V2' in the table. The original version V1 seemed to work better, but if fact a lot of the comparisons were very close. Each cell value indicates the best number of clusters with the related accuracy percentage and then which frequency grid version was used.

With relation to other heuristics, a value above 90% is considered to be acceptable for unsupervised clustering. That percentage level could be achieved for a single run, for example, 97% accuracy was possible with maybe twice as many clusters. In general, the clustering result fell a bit below that level. Adding the final clustering stage does not seem to improve the accuracy by very much, but it does appear to reduce the number of clusters, which suggests that it has helped to consolidate the result. This means that it can get closer to the desired number of clusters through the frequency grid stages. What is of interest is the difference between the first random column 3 and the ensemble versions, where the ensemble consistently outperforms the single clustering run. Comparing with the ordered dataset results of column 2, there is now a much smaller difference and so the ensemble process has compensated for the loss of accuracy in the frequency grid. The whole process however is quick and easy to use and even if the frequency grid is not the strongest algorithm for a particular problem, using the ensemble framework has consistently improved the results.

|  | O:B | O:B,F | R:B,F | R:B,F,E | R:B,F,E,H |
|---|---|---|---|---|---|
| Iris (150-3) | 4-68% | 5-96.5% V1 | 5-61% V1 | 6-87.5% V1 | 4-89.5% V1 |
| Wine (178-3) | 4-93% | 4-97% V1 | 11-62% V1 | 6-86% V1 | 5-91.5% V1 |
| Zoo (101-7) | 12-79% | 8-92% V1 | 8-52.5% V1 | 12-83% V1 | 8-88% V1 |
| Hayes-Roth (132-3) | 5-44.5% | 4-42% V1 | 4-44% V2 | 9-51.5% V1 | 4-47.5% V1 |
| Heart Disease Cleveland (303-5) | 8-57.5% | 6-56% V1 | 6-55% V1 | 8-58% V2 | 6-58% V2 |
| Sonar (208-2) | 3-54% | 4-59% V1 | 3-55% V1 | 5-62% V1 | 6-61% V2 |
| Wheat Seeds (210-3) | 5-89.5% | 5-88.5% V1 | 5-60.5% V1 | 5-85.5% V2 | 5-87.5% V1 |
| **Average Cluster Error** | 2.14 | 1.43 | 2.29 | 3.57 | 1.71 |
| **Average Accuracy Percentage** | 69.5% | 76% | 55.5% | 73.5% | 74.5% |

Table 2. Comparison of clustering methods: O: Ordered dataset, R: Random dataset, B: Brain-Inspired Closest nodes, F: Frequency Grid, E: Ensemble, H: Hierarchy.

## 7 Conclusions

This paper describes an unsupervised clustering approach that can then be corrected through a teaching process. The teaching phase however may allow the system to infer other classifications by itself and therefore reduce the time required to learn each data row. Data rows are firstly assigned to a classifier in an unsupervised manner, which represents them through a centroid. If there are errors, then further layers can create new centroids for subsets of the classifier rows. These centroids therefore define paths through the classifier branches and guide an input to its closest match. There is also a repeating process of re-assigning all data rows to the closest centroids, then re-calculating the centroids, to re-align the weight values with the current cluster sets. Splitting and re-combining clusters is another option. For this system, it would probably be better to have more smaller clusters that are more accurate. This is because the supervised part can provide additional help not available for the unsupervised clustering, but the unsupervised part still needs to make a reasonable attempt at producing accurate estimates. This paper explains the theory of the process and has described some unsupervised results only. It will be difficult to implement the whole system, as there are variations on what the best procedures might be.

As part of the ensemble learning, a final clustering stage does give the prospect of a hierarchical system that might increase its accuracy at each level. The clustering process would repeat and be fed input from the previous level. Any increase in accuracy would be attributed to the ensemble process and so the frequency grid heuristic could be replaced by another heuristic and results could still be aggregated together. So that is really interesting for a modular system such as the human brain, for example. A new teaching stage is proposed but has not been tested yet. It is interesting with respect to Explainable AI, which is important for improving trust in the the system. For one thing, the reduced training time and the ability to infer from another cluster's update would make the system more idependent. It is then also able to reason about how confident it is in its decision. If there are no links from an ensemble part to the hierarchy part, for example, then the system can say that it does not know for certain that the ensemble part is correct, or that the system should try to find out more about that section of data.

It is probably the case that unsupervised clustering cannot be as accurate as supervised and that is probably a good thing. On the one hand, the supervised clustering is making use of a lot of other information that has already determined what the correct clusters are, probably input by a human user. On the other hand, the human must have made these decisions at some stage through an unsupervised process and so if enough information is available, the computer system should be able to do it as well. The author still thinks that if AI becomes intelligent enough to take over, there is no reason to think that it will be evil, but is more likely to do good. However, the 'paper clips' scenario and the consequences of that is a bit clearer to him now. If a system is able to self-organise accurately enough with just some raw data, then it does not have any understanding outside of that and so it might in fact make bad decisions while having good intentions. So that could make the system more dangerous. Then again, if it ever reached a human level of intelligence, it would surely be able to learn when it had made a mistake and be able to correct it.